\documentclass[conference]{IEEEtran}

\usepackage{cite}
\usepackage{color}
\usepackage[stable]{footmisc}
\usepackage{siunitx}
\usepackage{authblk}
\usepackage[center]{subfigure}
\usepackage{verbatim}
\usepackage[super]{nth}
\usepackage{comment}
\usepackage{afterpage,lipsum}

\usepackage{amsfonts}
\usepackage[colorinlistoftodos]{todonotes}
\usepackage[linesnumbered, lined, commentsnumbered, ruled,vlined, longend, noend]{algorithm2e}
\usepackage{algpseudocode}
\usepackage{mathtools}
\usepackage{dirtytalk}
\usepackage{float}
\usepackage{times,amsmath,epsfig,amssymb,cite,bm,float,epstopdf,graphicx,graphics,amsthm}
\usepackage{forloop}
\usepackage{enumitem}
\newcounter{ct}

\newcommand{\cmmnt}[1]{\ignorespaces}

\usepackage{fancyhdr,lipsum}
\fancypagestyle{mahmood}{%
   \fancyhf{} 
   
   \fancyhead[C]{Accepted paper for publication in Globecom 2024. You can use this material personally. Reprinting or republishing this material for the purpose of advertising or promotion, creating new collective works, reselling or redistributing to servers or lists, or using any copyrighted component in other works must adhere to IEEE policy. The DOI will be supplied as soon as it becomes available.}

}%
\makeatletter
\let\ps@IEEEtitlepagestyle\ps@mahmood
\makeatother
\begin{document}

\title {Multi-Model based Federated Learning Against Model Poisoning Attack: A Deep Learning Based Model Selection for MEC Systems}

\author{Somayeh Kianpisheh\textsuperscript{1}}
\author{Chafika Benza\"{i}d\textsuperscript{1}}
\author{Tarik Taleb\textsuperscript{2}}
\vspace{-6cm}
\affil{\textsuperscript{1}\textit{Centre for Wireless Communications, University of Oulu, Finland}, \textsuperscript{2} \textit{Ruhr University Bochum, Germany} \\
\textit{Emails: somayeh.kianpisheh@oulu.fi, chafika.benzaid@oulu.fi, tarik.taleb@rub.de}}
\vspace{-6cm}
\maketitle
\begin{abstract}
Federated Learning (FL) enables training of a global model from distributed data, while preserving data privacy. However, the singular-model based operation of FL is open with uploading poisoned models compatible with the global model structure and can be exploited as a vulnerability to conduct model poisoning attacks. This paper proposes a multi-model based FL as a proactive mechanism to enhance the opportunity of model poisoning attack mitigation\cmmnt{, even for attacks that can deceive detection mechanisms}. A master model is trained by a set of slave models. To enhance the opportunity of attack mitigation, the structure of client models dynamically change within learning epochs, and the supporter FL protocol is provided. For a MEC system, the model selection problem is modeled as an optimization to minimize loss and recognition time, while meeting a robustness confidence. In adaption with dynamic network condition, a deep reinforcement learning based model selection is proposed. For a DDoS attack detection scenario, results illustrate a competitive accuracy gain under poisoning attack with the scenario that the system is without attack, and also a potential of recognition time improvement.   
\end{abstract}

\IEEEpeerreviewmaketitle
\textbf{Index Terms-} federated learning, poisoning attack, MEC, deep reinforcement learning.
\vspace{-0.3cm}
\section{INTRODUCTION}
Federated Learning (FL) provides collaboration opportunities for devices in training a global model from locally distributed data while preserving the data privacy by sharing model parameters instead of raw data \cite{mcmahan2017communication}. The learning is performed through some epochs. At each epoch, locally trained models of devices are transmitted to an aggregator server which aggregates the models to shape a global model for the next epoch of training. Utilizing a centralized server is inefficient due to single point of failure and intolerable latency in realtime scenarios \cite{zhou2022differentially}, \cite{kianpisheh2024collaborative}. FL in edge computing is an alternative where Multi-Access Edge Computing (MEC), collocated at base stations either aggregate the parameters from the devices in their district and return the results for a fast response, or partially aggregate the parameters and further transmit to a server for global aggregation \cite{zhou2022differentially}, \cite{kianpisheh2024collaborative}. Most of the FL studies in edge computing assume a secure FL protocol. 

A model poisoning attack exploits system vulnerabilities to control the global model's aggregation by generating poisoned local model updates to inject into the system so that the global model becomes either useless or less accurate. To make the FL robust, defense mechanisms mainly focus on mitigating model poisoning attack through outlier-detection mechanisms \cite{guerraoui2018hidden}, \cite{yin2018byzantine}, \cite{huang2021cost}, \cite{Jinhyun2024byzantine} or behavioural-based analysis \cite{al2023untargeted}, \cite{pan2020justinian}: The studies in \cite{guerraoui2018hidden}, \cite{yin2018byzantine} utilize robust aggregation rules in FL, that apply outlier detection methods to remove malicious models. The study in \cite{huang2021cost} provides a cost-efficient outlier detection method to isolate the attackers. The authors of \cite{chen2021fedequal} emphasize on performance of learning when applying an outlier detection. The authors of \cite{Jinhyun2024byzantine} make FL robust against Byzantine clients while preserving the privacy of individual users. The studies in \cite{al2023untargeted}, \cite{pan2020justinian} analyze the behaviours of the clients to determine unreliable clients and remove them in the aggregation phase. Either outlier-detection based mechanisms or mechanisms using behavioural-based analysis might not be efficient in recognition of new emerging model poisoning attack patterns, or attacks with complex behaviours. Also, the performance of outlier-detection methods reduces in heterogeneous environment where benign model might be considered as outlier\cite{chen2021fedequal}, \cite{baruch2019little}. In contrast to the reactive robustness mechanisms in the literature, this paper advocates a proactive approach to make the opportunity for poisoning lower, which can foster defense lines in addition to outlier-detection approaches, particularly for the attacks which can deceive the attack detection by intelligent model crafting. 

The poisoning attacks have a knowledge about global model structure so that they emulate a poisoned model compatible with global model. For example, \textit{A Little Is Enough} attack \cite{baruch2019little}, according to a Gaussian-based statistics deviates \textit{each dimension} of the model parameter from the \textit{mean} with a fraction of the \textit{standard deviation}. Model poisoning attack has been mathematically modeled in  \cite{al2023untargeted}. The attacker tries to deviate from global model with an arbitrary malicious model. Poisoned models compatible with global model structure, are uploaded. \textit{The poisoned model is calculated by updating global model with a learning rate based on the difference between the malicious target model and the global model}. 

The poisoning attacks use the singular-model based operation of FL as a vulnerability. They assume that all clients use the same model as the global model. Then, through applying an algorithm that follows their objectives in global model deviation, they emulate poisoned models with the same structure of the global model and inject them in learning process e.g., \cite{baruch2019little}, \cite{al2023untargeted}, \cite{guerraoui2018hidden}, \cite{huang2021cost}. This paper, proposes a multi-model based FL at which a master model or global model is trained by a set of slave models, letting dynamic change of client models within the learning epochs. 

Dynamic change of models that should be trained by clients can be either a barrier for attack operation or reduce the attack triggering chance. For example, \textit{A Little Is Enough} attack \cite{baruch2019little} \textit{may not be able to drive the mean and standard deviation for the global model dimensions}, when the clients will have heterogeneous models with different structures. Also, poisoning attacks e.g., \cite{baruch2019little}, \cite{al2023untargeted}, \cite{guerraoui2018hidden}, \cite{huang2021cost} construct a poisoned model compatible with the structure of the global model. \textit{When the client model is planned to be different from the global model, the mismatch between uploaded poisoned model and planed model, can be detected at aggregation time--providing the opportunity of attack detection and mitigation.} Emulation of client model is impossible if the communication protocol of FL be secure and the client model structure be dynamic and unknown for the attacker. Otherwise, random crafting of client model can be easily detected by detection mechanisms due to high deviation from normal behaviour. Sophisticated crafting of a client model \cmmnt{which dynamically changes and can be different than the global model,} demands designing new advanced attacks and introduces complexities. Advanced knowledge about e.g., dynamicity of the models, models' structures are required and new optimization for crafting should be derived to deviate global model in the desired direction, with crafting of heterogeneous, dynamic and different models from global model at client sides. To the best of our knowledge this sort of attack has not been investigated yet. Thus, multi-model based FL introduces potentials in dealing with poisoning attacks.  

This paper introduces \textbf{M}ulti-\textbf{M}odel based FL (MM-FL) as a proactive mechanism to reduce poisoning opportunity. An optimization is provided for model selection problem to minimize the loss and recognition time, while meeting a robustness confidence. To provide optimal model selection in large state/action spaces and under dynamic nature of wireless communication channels, a Deep Reinforcement Learning (DRL) based model selection is proposed. For a Distributed Denial of Service (DDoS) attack detection scenario, results illustrate improvement in accuracy and recognition time.

\vspace{-0.1cm}
\section{SYSTEM MODEL}
\textbf{Network:} A network consists of $N$ devices such as IoT and mobile devices, $M$ Base Stations (BSs) equipped with MEC servers, and a (central) cloud. Let $m^{th}$ $BS$ be represented with $BS_m$ with CPU frequency $f^{cmp}_m$. The available CPU cycle at the central cloud is $f^{cmp}_c$. We use symbol $u$ for a device, and its data with size $|D_{u}|$ is represented with $D_{u}=\{(x^{u}_{1},y^{u}_{1}),... (x^{u}_{|D_{u}|},y^{u}_{|D_{u}|})\}$, where $y$ is the label for input $x$. CPU frequency of the device $u$ is $f^{cmp}_u$ .\\
The model sharing can be done through MEC/cloud infrastructure, to construct a global model through FL. The aim of FL is to train a global learning model $M_c$ based on the distributed data in the devices, through some learning epochs. At each epoch of learning, a device performs learning on its local data to train a local model. The local models will evolve global model through FL process, based on which the required recognition is performed (e.g., attack detection).   

\textbf{Adversary Knowledge and Operation:} 
At any epoch of learning, the adversary can exploit vulnerabilities of devices and compromise them to inject poisoned models in learning process. The attacker does not have any control over aggregation process at MEC, nor over the protocol of the benign devices which follow a normal implementation of the protocol.

In consistent with the poisoned attacks in the literature, adversary has the knowledge about global model and emulates poisoned model(s) with the same structure as global model through applying an algorithm. It then uploads the malicious model(s) on behalf of the compromised device(s) when communicating with aggregator at MEC \cite{al2023untargeted}. The attack can be targeted or untargeted depending on the attacker objective, and our approach is applicable for both cases. In the case of targeted attack, the malicious model(s) are crafted so that the aggregated model over learning epochs approaches a targeted model \cite{al2023untargeted}. Otherwise, untargeted deviation strategy is employed in poisoning \cite{al2023untargeted}. 

\textbf{Federated Learning Objective:} The aim of federated learning is to train a global learning model, namely called master model based on the distributed data in the devices that take part in learning. Master model is trained through a set of slave models with the objective of making the FL robust with a predefined confidence level, while keeping the performance of FL in terms of accuracy and recognition time. 

\begin{figure}
\begin{center}
    \includegraphics[width=9cm, height=5cm]{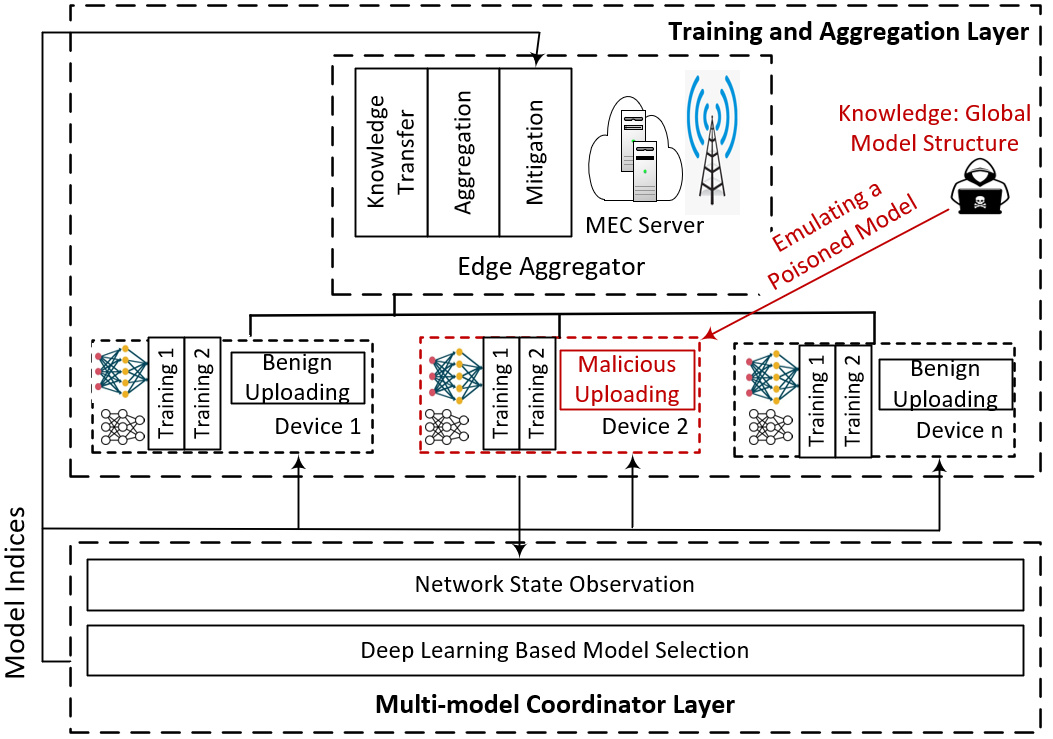}
    \vspace{-0.8cm}
     \caption{Multi-model FL architecture.}
    \vspace{-0.8cm}
\label{fig:arch}
\end{center}
\end{figure}
\vspace{-0.2cm}
\section{Multi-Model Based FL}
Fig. \ref{fig:arch} shows the architecture of multi-model based FL. Multiple models are used to train the main model, thereby having multiple training scripts at a device. At each learning epoch, network state is observed, according which a DRL-based model selection decides about the model that should be trained at each device. Then, the decision about models are announced to devices and they will train the model, they have been asked. After uploading model parameters, the edge aggregator will perform attack mitigation, aggregation, and knowledge transfer. The model parameters are downloaded for the next round. In the case of attack, the adversary that has knowledge about the global model structure, emulates poisoned models compatible with global model and upload them on behalf of the compromised devices. The attack can be mitigated before aggregation due to mismatching with the model plan which the multi-model coordinator arranged it.     
\vspace{-0.3cm}
\subsection{Fundamental} Let $M_c$ be the master model with weights $w_c$ and $M_s=\{M_s^1, ...M_s^l\}$ be the slave models with weights $\{w_s^1, ...w_s^l\}$. $M_c$ also can belong to the set of slave models, i.e., $M_c \in M_s$. \cmmnt{When $M_s=\{M_c\}$, we will have a conventional FL.} The main idea is to train master model using slave modes  according to an optimal model selection plan that operates as hindrance against Poisoning attack while maintaining the performance of FL.

Let $x_s^{j,u}(\tau)$ be the binary variable indicating the usage of model $M_s^j$ in federation of knowledge of device $u$ at epoch time $\tau$. The value of 1 indicates the usage of the model, while a value of 0 indicates not use of the model. As at every epoch of FL, only one model is selected for a device:

\vspace{-0.3cm}
{\footnotesize
\begin{equation}
\begin{aligned}
\forall \tau, u: 
\sum\limits_{j = 1}^{l}
     x_s^{j,u}(\tau) = 1. 
\end{aligned}
\vspace{-0.2cm}
\end{equation} 
}

We define the variable $\mathbb{T}_u(\tau)$ as the slave model that is selected to be used for the device $u$. We assume the associated weight is $\mathbb{W}_u$.    
The attack which emulates a poisoned global model on behalf of device $u$, can be detected at edge aggregator, due to a mismatch with the plan, if we have:  

\vspace{-0.2cm}
{\footnotesize
\begin{equation}
\begin{aligned}
\mathbb{T}_u(\tau) \neq M_c 
\end{aligned}
\vspace{-0.2cm}
\end{equation}
}
The aim is to secure the FL with a user defined confidence level, while maintaining the efficiency of FL performance. The diversity of models, will introduce hidden variables and will reduce the success of triggering poisoning attack. However, two issues will rise: First, new FL protocol should be designed to let the multi-model based learning process. Second, an optimization framework should be developed to decide about the models that are trained by devices, considering the dynamic nature of the network.

\vspace{-0.3cm}
\setlength{\textfloatsep}{0 cm}
\begin{algorithm}
\footnotesize
\caption{Knowledge Transfer}
\For{each BS $m$}
{
\For{each slave model $M_{e,m}^{s,j}$}
{
    \For{each instance in $D_{e,m}$}
    {
    input $\gets$ instance\\
    give input to model $M_{e,m}^{s,j}$\\
    label $\gets$ output neurons of $M_{e,m}^{s,j}$
    }
    $w_{e,m}(\tau) \gets$ GD optimization to minimize the loss function with the input for training as labeled $D_{e,m}$ 
}
}
\vspace{-0.1cm} 
\end{algorithm}

\vspace{-0.7cm}
\subsection{Multi-Model Based FL Protocol}
Multi-model based FL can train master model from slave model set. Master model can be included in the slave model set, since it can be the most efficient model in the context of recognition problem. The parameters of master model are found in learning process to minimize the global loss function in (4), where $f(w_c,x_s^{u}, y_s^{u})$ is the loss value over master model for the sample $s$ of data of user $u$.

\vspace{-0.2cm}
{\footnotesize
\begin{equation}
\begin{aligned}
\min\limits_{w_c} F =
\frac{1}{N}
\sum\limits_{u} 
     \frac{1}{|D_{u}|} \sum\limits_{s=1}^{|D_{u}|} f(w_c,x_s^{u}, y_s^{u}). 
\end{aligned}
\vspace{-0.2cm}
\end{equation} 
}

FL is performed through some epochs, indexed by $\tau$, until convergence or meeting of termination condition. BS $m$ trains a master model $M_{e,m}$, which has the same neural structure as $M_c$, from a set of slave models $M_{e,m}^s=\{M_{e,m}^{s,1}, ...M_{e,m}^{s,l}\}$ with the same structures of $M_s$. It also has an unsupervised data set of instances $D_{e,m}$ that is used for knowledge transfer. Each FL epoch, consists of the steps as below:

\begin{enumerate}     
    \item \textit{Local Training:} Each device employs a Gradient Decent (GD) based method using its local batch data to minimize the loss function over the planed model $\mathbb{T}_u(\tau)$. Then it transmits the local model parameters to its associated BS for the purpose of partial aggregation.
    \item  \textit{Attack Detection and Mitigation:} For each uploaded model from a device $u$ if the the parameters count are different than the planed model parameters count for that device ($\mathbb{W}_u(\tau)$), the model will be detected as malicious and will be excluded from the aggregation.
    \item \textit{Partial Aggregation at BSs:} For each slave model, the partial aggregation at BS $m$, is performed as a weighted average of received parameters of that model:
    
    \vspace{-0.3cm}
{\footnotesize
    \begin{equation}
    \begin{aligned}
    M_{e,m}^{s,j} = \frac{1}{K_{m,j}}=
         \sum\limits_{u=1}^{K_{m,j}} \mathcal{N}(|D_u|).w_u,
    \end{aligned}
    \vspace{-0.1cm}
    \end{equation} 
    }
    where $K_{m,j}$ is the number of devices under the coverage of $BS_i$, who use slave model $M_{e,m}^{s,j}$; and $|D_u|$ is the size of batch data of device $u$; and finally, $w_u$ is the weight of the trained model of device $u$. Symbol $\mathcal{N}(.)$ is the normalization operator.

    \item \textit{Knowledge Transfer To Master Model:} Each BS performs knowledge transfer step to transfer the knowledge of each slave model to the master model. First, the unsupervised data set is given to each slave model to be labeled. Then, the labeled data set is used to train and update the weights of the master model $w_{e,m}(\tau)$, based on a GD based optimization. Algorithm 1 is the pseudocode.
    \item \textit{Uploading of parameters:} Each BS transmits it's master model weights $w_{e,m}(\tau)$ to the cloud for the global aggregation.
    \item \textit{Cloud Aggregation:} The global model is constructed at central cloud as average of models of BSs and will be broadcasted to the BSs:

    \vspace{-0.2cm}
    {\footnotesize
        \begin{equation}
        \begin{aligned}
        M_{c}(\tau)= \frac{1}{M}
             \sum\limits_{m=1}^{M} M_{e,m}.
        \end{aligned}
        \vspace{-0.2cm}
        \end{equation} 
        }
                
    \item \textit{Broadcasting the Update Results:} Among the slave models in $M_{e,m}^{s}$, the model with the same structure of  $\mathbb{T}_u(\tau)$ will be transmitted to the user $u$.                    
    \end{enumerate}
FL epochs will be repeated, and finally the $M_c$ parameters will be broadcasted to the users at the final epoch or when the termination condition be met. \textit{Note that at step 6 after broadcasting of global model, the master model at BSs, and accordingly the devices that have uploaded the master model, will automatically be updated. This lets the master model update distribution even through training.}
\vspace{-0.2cm}
\section{Model Selection Optimization}
The 6G wireless model in \cite{fadlullah2020hcp} is used for communication between devices and the base stations. The transmission rate device $u$  communicates with BS $i$, is calculated by (6). $B_i$, $Pt_u$, $g_u$, $\eta$ are respectively, transmission bandwidth of the base station, transmission power of the device, channel gain of the device, and background noise power. Eq. (7) is the channel gain calculation. Here, $C_g$, $d_{u,i}$, $\alpha$ are respectively path loss fading coefficient, distance between device $u$ and base station $i$, and path loss exponent. \textit{Note that when devices move transmission rates will change due to distance variation.}   

\vspace{-0.2cm}
{\footnotesize
    \begin{equation}
        \begin{aligned}
            R_{u,i} = B_i\ln( 1 + \frac{Pt_u.g_u}{\eta} ),
        \end{aligned}
        \vspace{-0.2cm}
    \end{equation} 
        \begin{equation}
        \begin{aligned}
            g_u = C_g.d_{u,i}^{-\alpha},
        \end{aligned}
        \vspace{-0.2cm}
    \end{equation} 
}
The time cost of one epoch of FL is the time it takes a device receives the new updates of the model parameters. It includes the time slots allocated for local training, up/down-link parameter transmission, knowledge transfer, and the aggregation.  
\begin{itemize}
    \item \textit{Knowledge Transfer:} In BS $m$, the time for transferring knowledge to master model is calculated by:    

\vspace{-0.1cm}
{\footnotesize    
    \begin{equation}
    \begin{aligned}
        T_{knw}^{m} = \sum_{j} [\frac{|D_{e,m}|.f^{cmp}_{ds,j}}{f^{cmp}_{m}} + \frac{|D_{e,m}|.f^{cmp}_{ds,c}}{f^{cmp}_{m}}],
    \end{aligned}
    \vspace{-0.2cm}
    \end{equation}
}

    where $f^{cmp}_{ds,j}$, $f^{cmp}_{ds,c}$ are the number of required CPU cycles to label one unit of data by slave model $M_s^j$, and train master model $M_{e,m}$ with one unit of data, respectively. 
    
    \item {\textit{Aggregation:} The time for partial aggregation at BS $m$ includes: (a) the time it takes the parameters be transmitted from devices under the coverage area of that base station i.e., $R_m$  to the base station; (b) the aggregation operation time over the slave models:

    \vspace{-0.2cm}
{\footnotesize
    \begin{equation}
    \begin{aligned}
        T_{ag}^m = \max\limits_{u \in R_m} \frac{|\mathbb{W}_u|}{R_{u,i}} +        \sum_{j} \frac{X_s^j.|w_s^j|.f^{cmp}_w}{f^{cmp}_m},
    \end{aligned}
    \vspace{-0.1cm}
    \end{equation}
    }
    
    where $f^{cmp}_w$ is the number of required CPU cycles to aggregate one unit of data. $X_s^j$ is the number of devices under coverage of BS, that uploaded slave model $M_s^j$.\\
    The time for global aggregation at central cloud is calculated by (10). It includes the time for partial aggregation/knowledge transfer, at base stations, as well as the time takes for global aggregation of the parameters collected from base stations at the central cloud: 

    \vspace{-0.2cm}
{\footnotesize
    \begin{equation}
    \begin{aligned}
        T_{ag} = \max\limits_{m=1..M} (T_{ag}^m + T_{knw}^m + \frac{|w_c|}{R_{i,c}}) +
        \frac{M.|w_c|.f^{cmp}_w}{f^{cmp}_c}.
    \end{aligned}
    \end{equation}    
   }
   }    
    \item{
    \textit{Downlink Parameter Transmission:} The required time to download the parameters at device $u$ under the coverage of BS $m$ is calculated by (11), as the time required for parameters transmission from cloud to BS (with transmision rate $R_c$), and from the BS to the device.
    }  
    
        \vspace{-0.1cm}
    {\footnotesize
    \begin{equation}
    \begin{aligned}
        T_{down}(u) = \frac{|w_c|}{R_{c}} + \frac{|\mathbb{W}_u(\tau)|}{R_{m,u}}.
    \end{aligned}
    \end{equation}
    }
    
    \item{
    \textit{Local Training:} Local training time at device $u$ is calculated based on the computing capability of the device and the batch size. Eq. (12) is the calculation. Here, $f^{cmp}_s$ is number of required CPU cycles to train one sample of data over selected model.

    \vspace{-0.2cm}
{\footnotesize
    \begin{equation}
    \begin{aligned}
        T_{loc}(u) = \frac{|D_u|.f^{cmp}_s(\tau)}{f^{cmp}_u}.
        \end{aligned}
    \end{equation}
    }        
    }  
\end{itemize}
After the download of partially aggregated model, the device can perform the recognition. The recognition time in one epoch of FL, for device $u$, is calculated as below:

\vspace{-0.2cm}
{\footnotesize
\begin{equation}
    \begin{aligned}
        T_{Int}(u) =  \max_{u}K. T_{loc}(u) + T_{ag} + T_{down}(u) + \frac{f^{inf}_u}{f^{cmp}_u},
        \end{aligned}
\vspace{-0.1cm}
\end{equation}
}
where $K$ is the local training iterations before applying the other epoch of learning. $f^{inf}_u$ is the number of CPU cycles to perform the recognition for a sample of input features. The response time varies over epochs of learning. 

In a multi-model based FL, model selection for devices impacts the training and communication time, as well as the accuracy (See Eq. (9), (11), (12)). Straggler devices with poor transmission channel can prolong the learning process \cite{schlegel2023codedpaddedfl}. Selecting a smaller model than the master model, under poor transmission rates can speed up training, and compensate the latency of transmission, as well as reduce the transmission load--all speeding up the learning process. Furthermore, the security is enhanced since a different model than the global model can be utilized for training and poisoning attack can be mitigated at aggregation time (See Fig. 1). However, in the cases that global model be more efficient than slave models, there can be accuracy-reduction. On the other hand, in high quality of communication status, selecting global model in the case that it is more efficient than smaller model can enhance the accuracy, however the robustness reduces. Indeed, appropriate tradeoffs are required and optimization should be performed for the optimal selection.\\      
The slave models are assigned to the users, to minimize the loss function and recognition time over all users, over all epochs of learning. $\alpha$ and $\beta$ make the loss and time values in the same scale. Besides the already defined constraints, constrain (15) ensures the robustness against poisoning attack. $1.(B)$ is 1 if condition $B$ holds. To reduce the potential for poisoning, it ensures that the ratio of global model selection be less than a threshold value $T_{max} \in [0, 1]$. If $T_{max}=0$, poisoning attacks that emulate global model can not be triggered. However, the accuracy will be reduced if the master model be the most accurate model.   

\vspace{-0.1cm}
{\footnotesize
\begin{equation}
\begin{aligned}
\min\limits_{x_s^{j,u}(\tau)} \sum_{\tau, u} \alpha.F_u(\tau) + \beta.T_{Int}[\tau](u)
\end{aligned}
\end{equation} 
sbj.
(1), (2), (4), (5), (6), (7), (8), (9), (10), (11), (12), (13), (14),
\begin{equation}
\begin{aligned}
\forall \tau: \sum\limits_{u} 1.(\mathbb{T}_u(\tau)=M_c) \leq N.T_{max}.
\end{aligned}
\end{equation} 
}

\begin{figure}[t]
    \subfigure{
    \includegraphics[width=5.4cm, height=3cm]{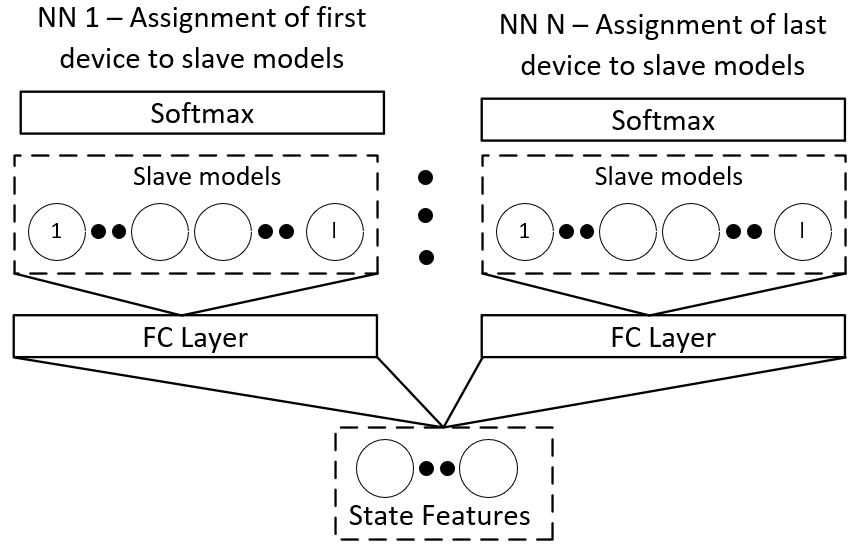}
    \label{fig:acc}
    }    
    \vspace{-0.4cm}
    \subfigure{
    \includegraphics[width=2.8cm, height=3 cm]{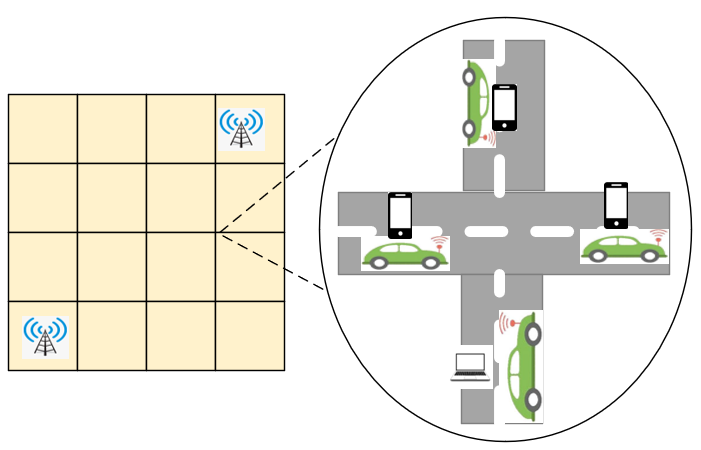}
     \label{fig:cost}
    }    
    \vspace{-0.3cm}
    \caption{(a) Policy networks. (b) Simulation grid.}   
\label{fig:NN}
\end{figure}

\vspace{-0.3cm}
\section{DRL-based Model Selection}
The search space order is exponential with dynamic channel communication. MDP and RL, advocated for solutions adoptable with dynamic network situations, can be used for formulation, since: (i) Function (14) has the memoryless property, as it can be calculated as the sum of loss and recognition time values in devices, in the current interval and the function value in the previous interval; (ii) Considering the parameters determining the current state e.g., transmission rates, association of slave models to devices, every action that is performed by the agent ends to a new state transition, that only depends on the current state. (iii) The function (14) is in the form of accumulated rewards. \cmmnt{Use of Reinforcement Learning (RL) in dynamic situation of network motivates its usage.} Through an iterative process of observing the state, choosing an action, and receiving a reward, the state-action Q-values are estimated by Bellman equation \cite{mnih2015human}. The high dimension of the states and the dynamicity in state transitions makes observing all states and actions in training impossible, thereby inefficiency of conventional RL. To deal with this problem, we adapt DRL \cite{mnih2015human}, that generalizes experienced states/actions to non-observed ones through a neural network-based approximation of Q-values.

\noindent \textbf{MDP Elements:} MDP Elements include:\\
\textit{State}: The features representing the state of network at time step $\tau$, are as below:
\begin{itemize}
    \vspace{-0.1cm}
    \item The transmission rates represented with matrix $\mathbb{R}_{d,e}(\tau)$, at which the entry at row $i$ and column $j$ is the transmission rate between device $i$ and BS $j$, at epoch $\tau$. Available bandwidth, distance of devices from BSs, channel gain varies the transmission rates. 
    \item The current slave model distribution denoted by vector $\mathbb{X}(\tau)$, at which the entry at row $j$ and column $u$ is the assignment of slave model $M_s^j$ to device $u$ at learning epoch $\tau$ (values of $x_s^{j,u}(\tau)$)   
\end{itemize}
\textit{Actions:} The action is decision about the assignment of devices to slave models (values of $x_s^{j,u}$ for the next epoch of learning).
\textit{Reward:} In the case of violation of constraints (1), (15), reward is 0. To ensure, optimizing the objective function the reward function is calculated as: 

\vspace{-0.3cm}
{\footnotesize
\begin{equation}
\begin{aligned}
\mathbb{R}(s(\tau),a(\tau))=\sum_{u} \alpha.F_u(\tau) + \beta.T_{Int}[\tau](u)
\end{aligned}
\vspace{-0.2cm}
\end{equation} 
}

\noindent \textbf{Training:} Using $N$ policy networks, the decision policy is derived by training them. Each Neural Network (NN) represents the assignment of a device to the slave models. The input neurons are the state features. There is a Fully-Connected layer, with Softmax activation function. The neuron $i$ in the output layer of NN $u$, indicates the probability of assigning the slave model $M_s^i$ to the device $u$ (See Fig. \ref{fig:NN}.a).\\
Each episode consists of a run of FL within several epochs of learning. There are variation in network communication status, location of devices, and compromised devices within episodes. Training is done through two steps performed at every epoch of learning: (i) \textit{Exploration:} According to $\epsilon-$greedy policy, with a $1-\epsilon$ probability, a random assignment of slave models to devices is selected. Otherwise, the current state features are given as input to the NNs. For each device, the slave model, with the highest probability at output layer, will be assigned to the device. (ii) \textit{Updating the weights:} After the assignment strategy, reward is calculated by (16), accordingly the NNs' weights are updated by Gradient Descent (GD) method, and using Bellman equation \cite{mnih2015human}.  

\begin{figure*}[!h]
\begin{center}
    \begin{minipage}{5.5cm}
    \begin{center}
        \includegraphics[width=5.4cm, height=3.6cm]{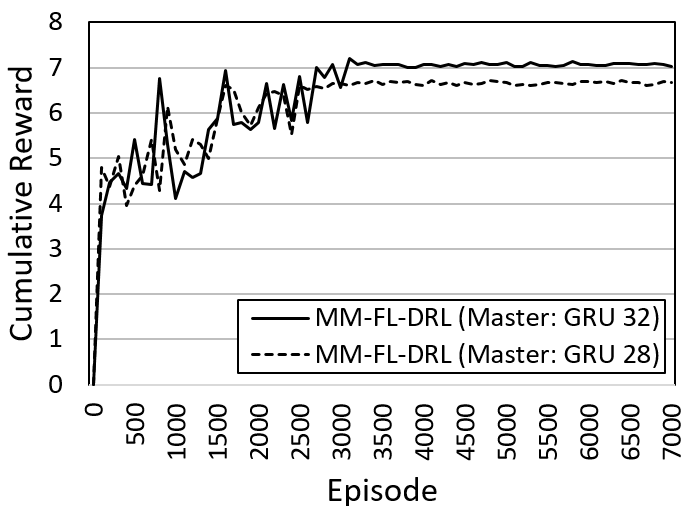}
        \label{fig:reward}
        \vspace{-0.6cm}
        \caption{Cumulative reward.}
    \end{center}
    \end{minipage}
    \begin{minipage}{5.8cm}
    \begin{center}
        \includegraphics[width=5.8cm, height=3.6cm]{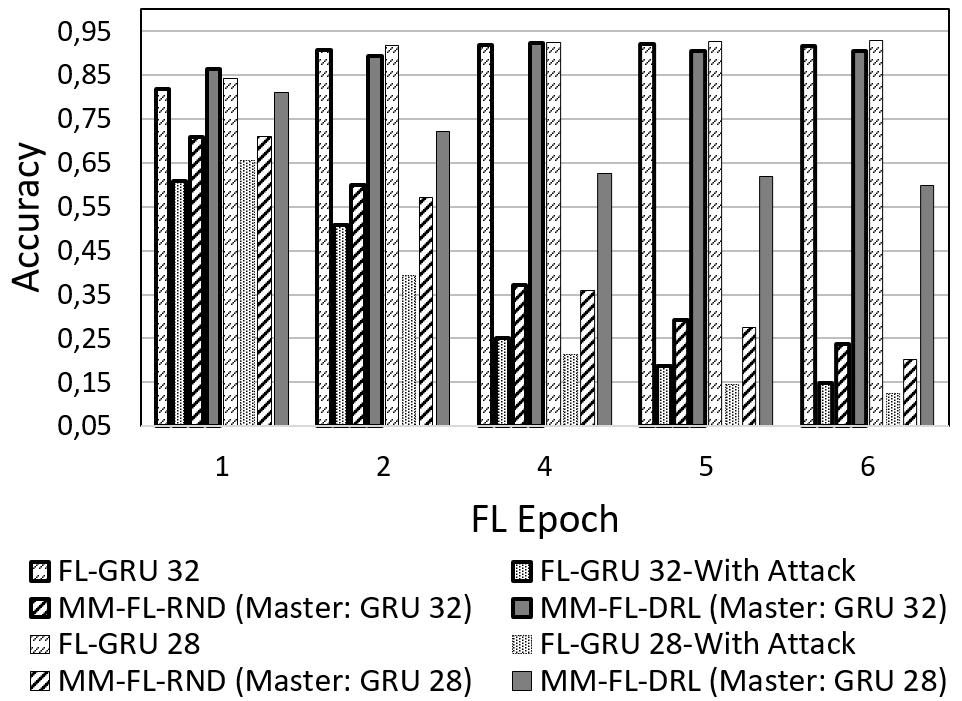}
        \label{fig:acc}
        \vspace{-0.6cm}
        \caption{DDoS attack detection accuracy.}
        \end{center}
    \end{minipage}
    \begin{minipage}{5.8cm}
    \begin{center}
        \includegraphics[width=5.8cm, height=3.6cm]{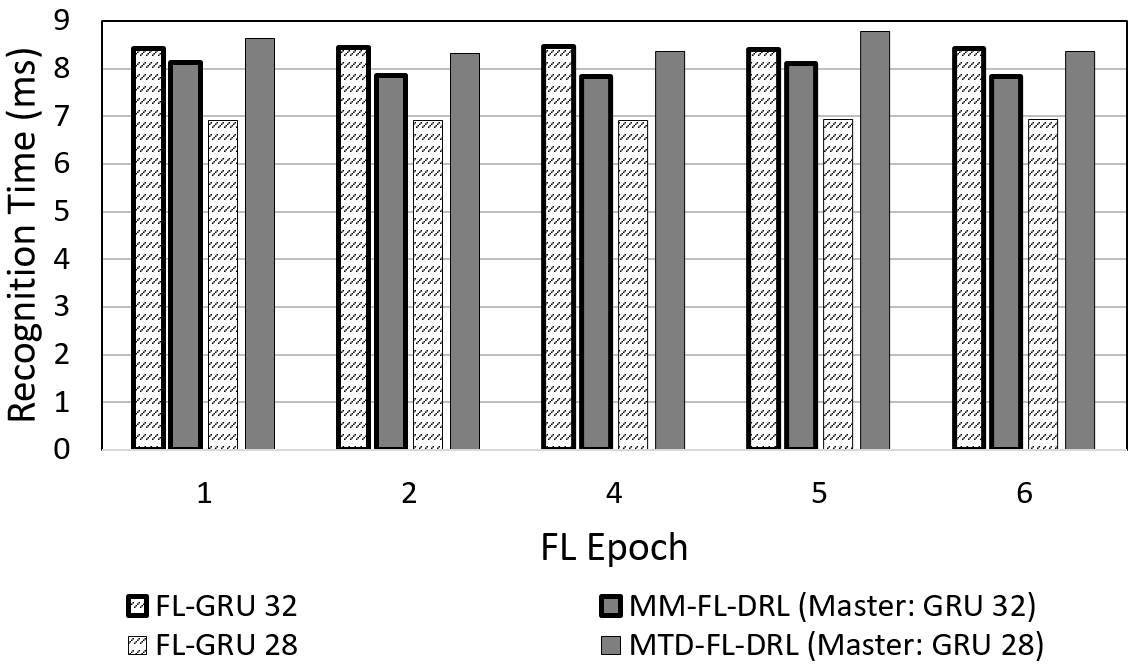}
        \label{fig:T}
        \vspace{-0.4cm}
        \caption{Recognition time.}
    \end{center}
    \end{minipage}
\end{center}
\vspace{-1cm}
\end{figure*} 
\section{Experimental Results}
A scenario that 10 devices that are moving by vehicles, will collaborate in DDoS attack detection, is considered. The advantage of applying FL in DDoS attack detection in terms of enhancing accuracy through sharing attack detection models in comparison with individual (local) learning approaches have been illustrated in many recent studies e.g., \cite{kianpisheh2024collaborative}, \cite{li2021fleam}. The study in \cite{kianpisheh2024collaborative}, extensively discusses the FL benefits for DDoS attack detection. We used CICDDoS 2019 data set~\cite{sharafaldin2019developing} comprising realistic traffic to abstract the communications for legitimate and DDoS attack traffic through protocols e.g., HTTP, FTP. We have randomly distributed 9000 instances composed of UDPLag and SYN DDoS attacks, among devices for training, as well as 3000 instances for the purpose of test. The dataset provides 87 IP flow features e.g., source/destination IP addresses/ports, protocols, flow packet statistics, flag-related information etc., which we utilize them for the attack detection. 

We assume a $4 \times4$ bidirectional grid environment with $100~ m$ width for each grid cell, where grid lines are bidirectional roads (Fig. \ref{fig:NN}.b). Mobility traces of vehicles have been generated by SUMO simulator~\cite{SUMO}. Manhattan mobility model in urban areas~\cite{SUMO} with probability of 0.5 for moving straight and 0.25 for moving right/lef at conjunctions, is used for mobility of vehicles. The mean speed of vehicles are 45 km/h. \textit{The mobility of vehicles will cause dynamicity in transmission rates.} We used GRU as we explained it with details in \cite{kianpisheh2024collaborative} for attack detection. The feature matrix for the packets in a flow is arranged as rows of patterns. The occurrence probability for each pattern is calculated as a function of previous observations using a GRU. A flow (including 10 packets) is malicious if the ratio of the malicious packets in that flow is larger than a threshold (0.7). TensorFlow and Keras are used to implement the GRUs/DRL.

Without loss of generality, the learning collaboration is only performed in the level of edge computing. Two BSs equipped with MEC-servers with respectively CPU frequencies of 3.2 and 2.6 GHz \cite{lu2020low} are located in the locations $[50, 50], [350, 350]$ to provide edge coverage for devices. Each BS has 2400 random unsupervised instances from CICDDOS data set for knowledge transfer. The coverage radius of base stations are 300 m, and their transmission bandwidth are respectively, 28 and 30 MHz \cite{kianpisheh2024collaborative}. The CPU frequency of devices are randomly chosen in the range of 1.9 up to 2.4 GHz. The transmission power of BSs and devices are respectively 34 db and 23 db \cite{lu2020low}. Path loss exponent is 5 and back ground noise power is -174 db.m \cite{lu2020low}.  

There are more explorations at early iterations of DRL, while the exploitation gradually increases up to the greedy selection of 98\% at the last episode. For GRUs, the learning rates 0.07 for GD optimization operated efficiently.  $LearningRateSchedule$ package of Keras reduces the learning rate within training for the purpose of convergence. The out-layer of the GRU is a neuron to predict the occurrence probability of a packet. See \cite{kianpisheh2024collaborative} for details. Discount rate of 0.1 and ADAM optimization in DRL also operated efficiently.

In the rest, GRU with 28 and 32 neurons in the hidden layer are represented with GRU 28 and GRU 32, respectively. Various scenarios have been considered: Two FL scenarios when GRU 28 and GRU 32 are applied for DDoS attack detection. Each FL is applied in a system without attack (FL-GRU 28, FL-GRU 32), and a system with poisoning attack (FL-GRU 28-With Attack, FL-GRU 32-With Attack). In MM-FL the slave models set include GRU 28 and GRU 32. Two scenarios of MM-FL are considered: MM-FL with master model as GRU 28 and MM-FL with master model as GRU 32. We have MM-FL-DRL which utilizes DRL in model selection, and MM-FL-RND which uses a random selection of slave models. At each episode, the adversary compromises 3 to 5 random devices and emulates Malicious Local Models (MLMs) with the global model structure, for updating  using the method in \cite{al2023untargeted}, as explained in Introduction, with  random target malicious models and random learning rate in (0.25, 0.35). The results are average over 50 runs of test, $T_{max}=0.6$. 

Fig. 3 illustrates the cumulative reward gain within episodes in MM-FL. The cumulative reward has increased up to range of 6.5 to 7 and become stable around episode 3200, with a dominance in reward with the case that GRU 32 is master model. Since, the scenario that GRU 32 is master has gained better performance in terms of accuracy and recognition time. 

Fig. 4 shows the accuracy of DDoS attack detection. The bars with bold boarders are for GRU 32 scenarios, and the rest are for GRU 28 scenarios. For GRU 32 scenarios, the accuracy considerably decreases after poisoning, and this decrease becomes the worse at the late epochs of FL due to more involvement of MLMs in aggregation e.g., drop from 0.92 to 0.15 at epoch 6. MM-FL-RND has increased the accuracy slightly e.g., for 0.09 in epoch 6. MM-FL-DRL has increased the accuracy considerably and in competitive with the scenario of FL without attack, due to optimization of loss function in model selection and  selecting atleast 60\% of the local models different than the main model, according which MLMs can be detected and mitigated. Similar results has been achieved for GRU 28 scenarios. However, MM-FL-DRL (Master: GPU 32) operates better than MM-FL-DRL (Master: GPU 28), which can illustrate the more efficiency of knowledge transfer from GRU 28 to GRU 32, in comparison with the reverse in MM-FL-DRL (Master: GPU 28). This is an application dependent issue.          

Fig. 5 shows the recognition time. For GRU 32 scenario, the recognition time reduces up to 0.6 ms, after multi-model FL application with GRU 32 as master. Since, in this scenario, MM-FL-DRL uses GPU 28 for atleast 60\% of devices which demands less training and transmision time than GRU 32. However, application of MM-FL with GPU 28 as master, increases recognition time in comparison with FL with GRU 28 due to employing larger model of GRU 32 in training.  

\vspace{-0.2cm}
\section{Conclusion} \label{sec:sec5}
This paper proposes a multi-model based FL as a proactive mechanism to enhance the opportunity of model poisoning attack detection and mitigation. A protocol at which a master model is trained by a set of slave models, is explained. For federation over a MEC system, the model selection problem is modeled as an optimization and a DRL-based model selection method adaptable with dynamic network conditions is provided. For a DDoS attack detection scenario, results illustrate a competitive accuracy with the scenario without attack. With a smaller model employment in training recognition time reduces. 
\vspace{-0.3cm}
\bibliographystyle{IEEEtran}
\bibliography{ref.bib}
\end{document}